%% file: main.tex
\documentclass{article}
\usepackage[preprint]{neurips_2025}

\usepackage{natbib} 
\usepackage[colorlinks=true, citecolor=blue, linkcolor=blue, urlcolor=blue, anchorcolor=blue]{hyperref}
\usepackage{graphicx}
\usepackage{subcaption}
\usepackage[utf8]{inputenc} 
\usepackage[T1]{fontenc}    
\usepackage{breakurl}     
\usepackage{listings}
\usepackage{caption}
\usepackage{booktabs}       
\usepackage{listings}
\usepackage{placeins}
\usepackage{amsfonts}       
\usepackage{amsmath}
\usepackage{amsthm}
\usepackage{xurl}
\usepackage{amssymb}
\usepackage{nicefrac}       
\usepackage{microtype}      
\usepackage[font=small,labelfont=bf]{caption}
\usepackage{enumitem}
\usepackage{wrapfig}
\usepackage{listings}
\usepackage{caption}
\usepackage{multirow}
\usepackage[most]{tcolorbox}
\usepackage[normalem]{ulem}
\usepackage{xspace}
\usepackage{float}
\usepackage{tabularx}
\usepackage{booktabs}
\usepackage[normalem]{ulem}
\usepackage{soul}   
\usepackage{svg}
\usepackage{array}
\usepackage{longtable}
\usepackage{adjustbox}
\usepackage{makecell}
\usepackage{pifont}
\usepackage{pdflscape}

\usepackage{xcolor}

\useunder{\uline}{\ul}{}

\definecolor{mygreybg}{gray}{0.95}

\sethlcolor{mygreybg}

\title{Voxtral Realtime}

\begin{document}
\maketitle
\vspace{-0.1in}
\begin{center}
\vspace{-45pt}
\centering
\includegraphics[width=0.8\linewidth,keepaspectratio]{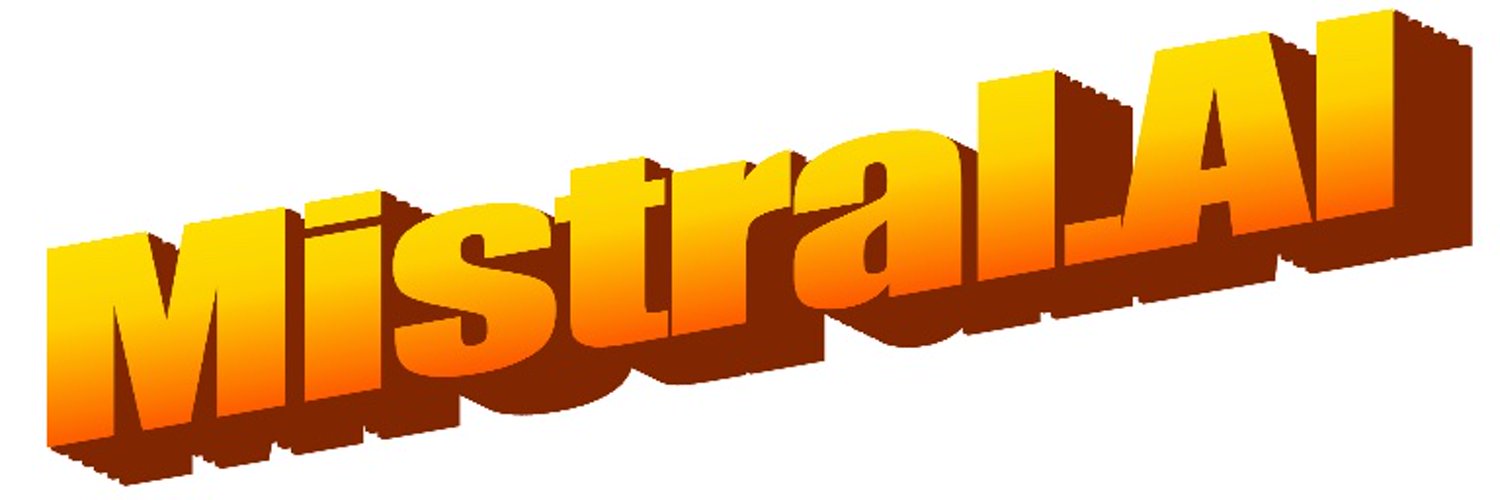}
\end{center}

\begin{abstract}
\input{abstract.tex}
\end{abstract}
\begin{center}

\begin{tabular}{@{} l l @{}}
\small{\textbf{Webpage:}}  & \scriptsize{\url{https://mistral.ai/news/voxtral-transcribe-2}} \\
{\small{\textbf{Model weights:}}}  & \scriptsize{\url{https://huggingface.co/mistralai/Voxtral-Mini-4B-Realtime-2602}} \\
\end{tabular}
\end{center}

\vspace{-0.3cm}


\begin{figure}[h]
    \centering
    \includegraphics[width=0.85\textwidth]{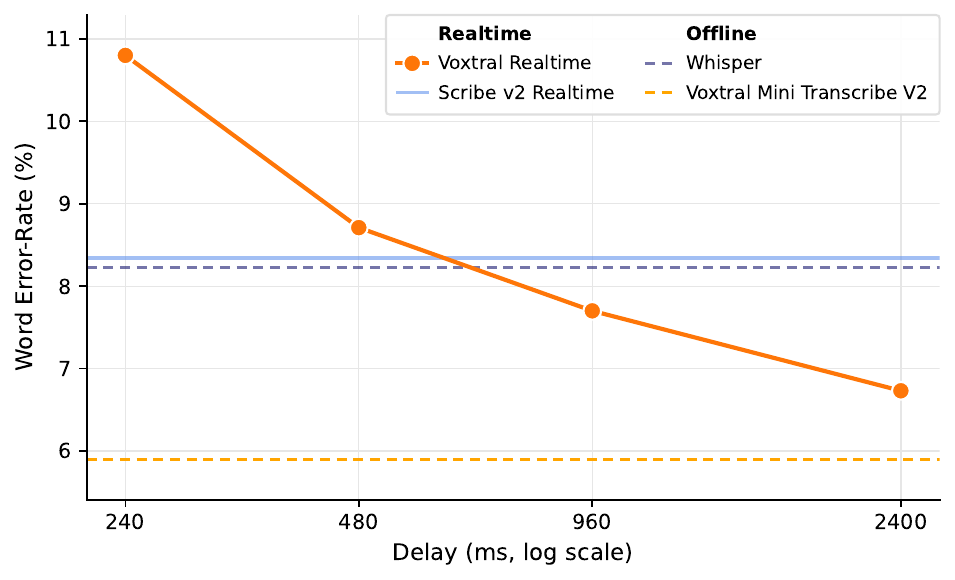}
    \caption{\textbf{Voxtral Realtime approaches offline accuracy at sub-second latency.} Macro-average word error-rate (WER) vs. delay on the FLEURS multilingual benchmark for realtime and offline models. Lower is better. At 480\,ms delay, Voxtral Realtime is competitive with Scribe v2 Realtime, the leading realtime API model, as well as Whisper, the most popular open-source offline model. It surpasses both baselines at 960\,ms delay, approaching the performance of Voxtral Mini Transcribe V2, a state-of-the-art offline transcription model.}
    \label{fig:fleurs-latency}
\end{figure}

\section{Introduction}
\input{introduction.tex}


\section{Modeling}
\input{modeling.tex}

\section{Training} \label{sec:training}
\input{training.tex}

\section{Inference and Serving in vLLM}
\input{inference.tex}

\section{Results}
\input{results.tex}

\section{Analysis}
\input{ablations.tex}

\section{Conclusion}
\input{conclusion}

\subsection*{Core contributors}

Alexander H. Liu, Andy Ehrenberg, Andy Lo, Chen-Yo Sun, Guillaume Lample, Jean-Malo Delignon, Khyathi Raghavi Chandu, Patrick von Platen, Pavankumar Reddy Muddireddy, Rohin Arora, Sanchit Gandhi, Sandeep Subramanian, Soham Ghosh, Srijan Mishra.

\subsection*{Contributors}

Abhinav Rastogi, Adrien Sadé, Alan Jeffares, Albert Jiang, Alexandre Cahill, Alexandre Gavaudan, Alexandre Sablayrolles, Amélie Héliou, Amos You, Andrew Bai, Angele Lenglemetz, Anmol Agarwal, Anton Eliseev, Antonia Calvi, Arjun Majumdar, Avi Sooriyarachchi, Baptiste Bout, Baptiste Rozière, Baudouin De Monicault, Benjamin Tibi, Charlotte Cronjäger, Clémence Lanfranchi, Connor Chen, Corentin Barreau, Corentin Sautier, Cyprien Courtot, Darius Dabert, Diego de las Casas, Elizaveta Demyanenko, Elliot Chane-Sane, Enguerrand Paquin, Etienne Goffinet, Fabien Niel, Faruk Ahmed, Federico Baldassarre, Gabrielle Berrada, Gaëtan Ecrepont, Gauthier Guinet, Genevieve Hayes, Georgii Novikov, Giada Pistilli, Guillaume Kunsch, Guillaume Martin, Guillaume Raille, Gunjan Dhanuka, Gunshi Gupta, Han Zhou, Harshil Shah, Hope McGovern, Hugo Thimonier, Indraneel Mukherjee, Irene Zhang, Jaeyoung Kim, Jan Ludziejewski, Jason Rute, Joachim Studnia, John Harvill, Jonas Amar, Joséphine Delas, Josselin Somerville Roberts, Julien Tauran, Karmesh Yadav, Kartik Khandelwal, Kilian Tep, Kush Jain, Laurence Aitchison, Laurent Fainsin, Léonard Blier, Lingxiao Zhao, Louis Martin, Lucile Saulnier, Luyu Gao, Maarten Buyl, Manan Sharma, Margaret Jennings, Marie Pellat, Mark Prins, Martin Alexandre, Mathieu Poirée, Mathilde Guillaumin, Matthieu Dinot, Matthieu Futeral, Maxime Darrin, Maximilian Augustin, Mert Unsal, Mia Chiquier, Minh-Quang Pham, Nathan Grinsztajn, Neha Gupta, Olivier Bousquet, Olivier Duchenne, Patricia Wang, Paul Jacob, Paul Wambergue, Paula Kurylowicz, Philippe Pinel, Philomène Chagniot, Pierre Stock, Piotr Miłoś, Prateek Gupta, Pravesh Agrawal, Quentin Torroba, Ram Ramrakhya, Rishi Shah, Romain Sauvestre, Roman Soletskyi, Rosalie Millner, Rupert Menneer, Sagar Vaze, Samuel Barry, Samuel Humeau, Sean Cha, Shashwat Verma, Siddhant Waghjale, Siddharth Gandhi, Simon Lepage, Sumukh Aithal, Szymon Antoniak, Teven Le Scao, Théo Cachet, Theo Simon Sorg, Thibaut Lavril, Thomas Chabal, Thomas Foubert, Thomas Robert, Thomas Wang, Tim Lawson, Tom Bewley, Tom Edwards, Tyler Wang, Umar Jamil, Umberto Tomasini, Valeriia Nemychnikova, Van Phung, Vedant Nanda, Victor Jouault, Vincent Maladière, Virgile Richard, Vladislav Bataev, Wassim Bouaziz, Wen-Ding Li, William Havard, William Marshall, Xinghui Li, Xingran Guo, Xinyu Yang, Yannic Neuhaus, Yassine El Ouahidi, Yassir Bendou, Yihan Wang, Yimu Pan, Zaccharie Ramzi, Zhenlin Xu.

\subsection{Acknowledgements}

We would like to thank Joshua Deng, Yu Luo from Meta AI, and Nick Hill, Nicolò Lucchesi, Chen Zhang, Cyrus Leung, and Roger Wang from the vLLM team for their support and contributions in integrating Voxtral Realtime to the vLLM framework.

We are grateful to Salvatore Sanfilippo, Awni Hannun, Prince Canuma, TrevorS, Eustache Le Bihan and the open-source community for their contributions of Voxtral Realtime to additional frameworks.

\clearpage

\bibliography{ref}

\clearpage

\appendix
\input{appendix}

\end{document}

%% file: abstract.tex
We introduce Voxtral Realtime, a natively streaming automatic speech recognition model that matches offline transcription quality at sub-second latency. Unlike approaches that adapt offline models through chunking or sliding windows, Voxtral Realtime is trained end-to-end for streaming, with explicit alignment between audio and text streams. Our architecture builds on the Delayed Streams Modeling framework, introducing a new causal audio encoder and Ada RMS-Norm for improved delay conditioning. We scale pretraining to a large-scale dataset spanning 13 languages. At a delay of 480ms, Voxtral Realtime achieves performance on par with Whisper, the most widely deployed offline transcription system. We release the model weights under the Apache 2.0 license.

%% file: introduction.tex
Automatic speech recognition (ASR) systems achieve strong performance in offline settings \citep{radford2023whisper,liu2025voxtral}, where the entire audio input is available before transcription begins. However, many real-world applications—such as voice assistants, live captioning, and interactive speech interfaces—require transcriptions to be produced in real time while audio is streaming, under strict latency constraints. Bridging the gap between offline transcription quality and real-time streaming remains a central challenge in speech recognition \citep{graves2012rnnt,zeghidour2025streaming}.

A common approach to streaming adapts offline models by processing audio in short chunks as it arrives \citep{machacek2023streaming}. While effective at moderate latencies, this strategy has a fundamental limitation: offline models are typically trained with access to bidirectional acoustic context (and often full-sequence conditioning), whereas a streaming system must emit tokens before future audio is available. This training--inference mismatch becomes increasingly severe as latency is reduced, and often leads to degraded accuracy in low-delay regimes and out-of-distribution settings.

\looseness=-1 Native streaming architectures address this by reformulating the learning problem so that each output token is predicted using only past inputs and a bounded amount of lookahead, making latency a tunable constraint. This requires (i) an explicit alignment between input audio and output text (e.g., word- or frame-level alignments), and (ii) an architecture that processes new audio incrementally as it arrives. A common instantiation is a neural transducer (RNN-T) with a streaming encoder that limits right context through chunking, memory, and caching \citep{graves2012rnnt,Shi2020EmformerEM, chen2021developing, noroozi2024nemotron}. Delayed Streams Modeling (DSM) \citep{zeghidour2025streaming} follows the same alignment-based principle, but replaces the transducer with a decoder-only model over aligned audio and text streams, enabling simpler designs that leverage pre-trained language decoders. DSM approaches offline accuracy at high delay settings. However, achieving offline-level performance at sub-second latency—particularly in multilingual and multi-domain settings—has remained an open challenge.

We address this challenge by introducing \emph{Voxtral Realtime}, a 4B parameter natively streaming ASR model that supports 13 languages. 
Concretely, our contributions are:
\begin{itemize}
    \item A causal audio encoder trained from scratch with modern architectural choices (RMSNorm, SwiGLU, RoPE, sliding window attention).
    \item An adaptive RMS-Norm (Ada RMS-Norm) mechanism in the decoder, enabling a single model to operate at any delay that is a multiple of 80\,ms.
    \item Pretraining at scale on a large-scale dataset spanning 13 languages, enabling robust generalization across languages and domains.
\end{itemize}

\looseness=-1 At a delay of 480\,ms, Voxtral Realtime achieves performance competitive with Whisper \citep{radford2023whisper} and ElevenLabs Scribe v2 Realtime \citep{scribev22025}. At higher delay settings (e.g., 960\,ms), it matches or surpasses strong offline baselines such as Voxtral Mini Transcribe V2 on several English and multilingual benchmarks \citep{voxtral22026}. These results demonstrate that offline-level transcription quality can be achieved within a fully streaming framework at sub-second latency.

We release the resulting model as open weights under the Apache 2.0 license. The remainder of this report details the model architecture, training and inference methodology, and empirical evaluations that support these findings.

%% file: modeling.tex
\begin{figure}[t]
\centering
\includegraphics[width=\textwidth]{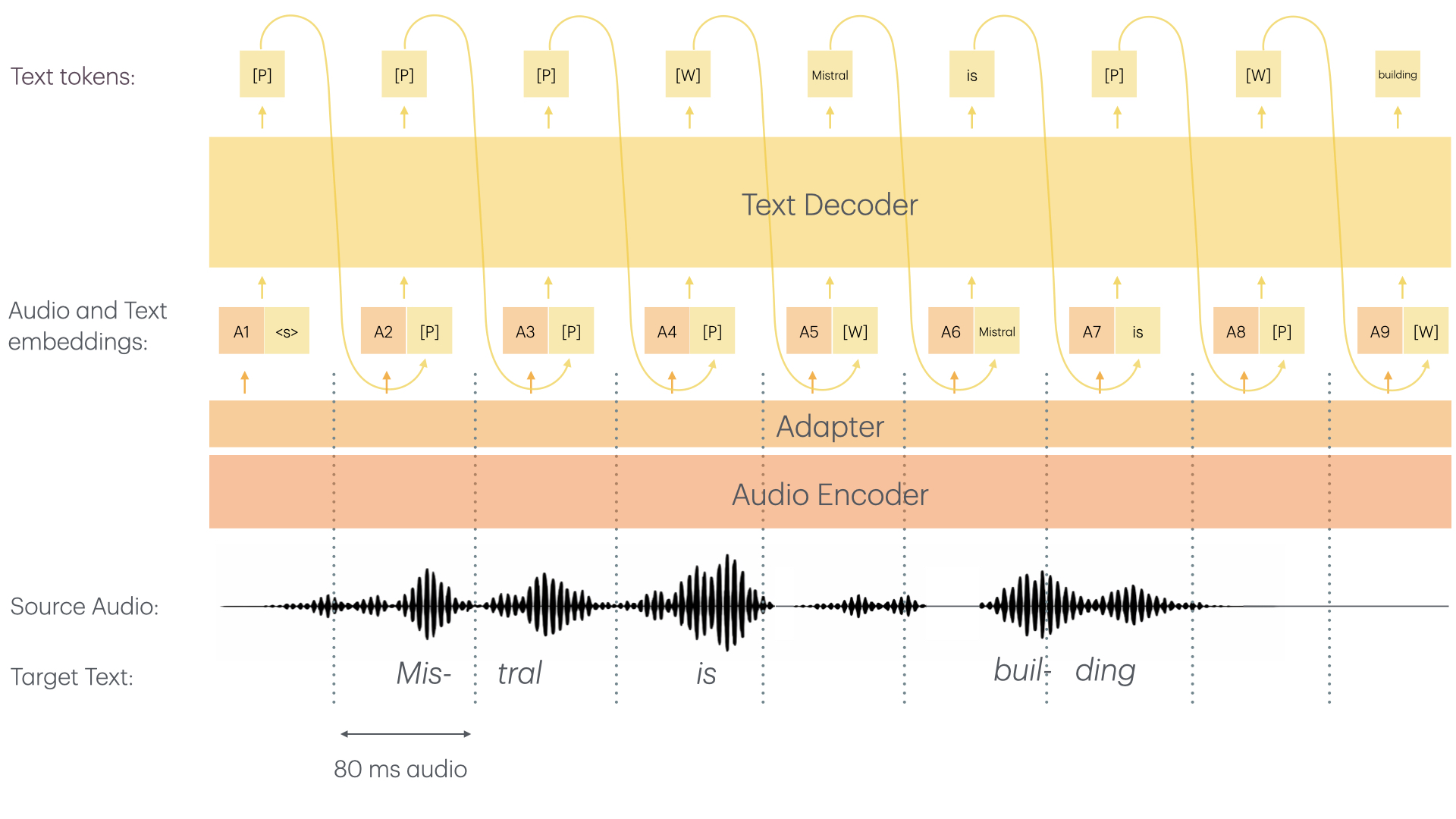}
\caption{
\label{fig:archi}
\textbf{Voxtral Realtime architecture and decoding scheme for a target delay $\tau=80$\,ms.} Voxtral Realtime consists of a causal audio encoder to embed the input audio stream, an MLP adapter layer to temporally downsample the audio embeddings, and a text decoder to auto-regressively generate the output text stream. The downsampled audio embeddings from the adapter and the embeddings of previously generated tokens have the same frame-rate of 12.5Hz, with each frame representing 80ms of audio. These are summed and processed by the text decoder, which predicts one token per frame. The decoder emits a padding token \texttt{[P]} while waiting for sufficient acoustic evidence. Once a word is acoustically complete and the target delay $\tau$ has elapsed, a word-boundary token \texttt{[W]} is emitted to initiate generation, followed by the corresponding subword tokens.
}
\end{figure}

Voxtral Realtime is a Transformer-based streaming ASR model that follows the stream-synchronous design of DSM. The model comprises (i) a causal audio encoder, (ii) a temporal adapter that downsamples encoder frames, and (iii) a Transformer decoder that generates text autoregressively. At each stream step, the decoder consumes a fused representation obtained by summing the current-step audio embedding with the embedding of the most recently generated text token. The overall architecture is summarized in Figure~\ref{fig:archi}, with model dimensions outlined in Table~\ref{tab:model-dims}.

\begin{table}[h]  
\centering          
\caption{
\textbf{Voxtral Realtime configuration}. For the decoder, we use grouped-query attention (GQA) \citep{ainslie2023gqa}; the number in parentheses indicates KV heads. Sliding window sizes are in frames (encoder) and tokens (decoder).
}
\label{tab:model-dims}                 
\begin{tabular}{lccccc}                   
\toprule            
\textbf{Component} & \textbf{Layers} & \textbf{Dimension} & \textbf{Heads} & \textbf{Sliding Window} & \textbf{Parameters} \\     
\midrule            
Audio Encoder & 32 & 1280 & 32 & 750 & 970M \\
Adapter MLP & 1 & 1280$\times$4 $\rightarrow$ 3072 & — & — & 25M \\
Language Decoder & 26 & 3072 & 32 (8 KV) & 8192 & 3.4B \\                  
\midrule            
\textbf{Total} & — & — & — & — & 4.4B \\     
\bottomrule         
\end{tabular}       
\end{table}

\subsection{Audio Encoder}

Audio encoders for offline ASR systems are typically trained with bidirectional attention \citep{baevski2020wav2vec20,radford2023whisper}, since the full audio signal is available. However, in real-time settings the encoder must produce representations causally, attending only to current and past inputs.

Therefore, we define a causal audio encoder architecture and train it from scratch. The waveform is converted to a log-Mel spectrogram \citep{davis1980comparison} with 128 Mel bins and a hop length of 160 samples (10\,ms at 16\,kHz). Features are processed by a causal convolutional stem with 2x temporal downsampling, followed by a stack of causal self-attention layers. The encoder emits one embedding every 20\,ms (50\,Hz).

\looseness=-1 The convolutional stem induces a finite history dependency: the output at step $t$ depends on the previous four input frames (two kernel-3 convolutions, including a strided downsampling layer). During streaming inference, we maintain a 4-frame history buffer to compute the current encoder state exactly.

\looseness=-1 In the Transformer backbone, we adopt RMSNorm, SwiGLU and RoPE—architectural choices that have been shown to improve training stability and downstream performance \citep{zhang2019rms,shazeer2020swiglu,su2023rope,touvron2023llama}. The self-attention uses a sliding window of 750 frames (15\,s at 50\,Hz) \citep{child2019sliding,beltagy2020long}, bounding memory while enabling unbounded streaming. The differences in relation to the Whisper encoder are summarized in Table~\ref{tab:archi}.

\begin{table}[t]
\centering
\caption{\textbf{Whisper vs. Voxtral Realtime encoder architectures.} Voxtral Realtime is a fully causal encoder that leverages modern architectural choices with a 750 frame sliding window.}
\label{tab:archi}
\begin{tabular}{lcccccc}
\toprule
\textbf{Model}   & \textbf{Attention} & \textbf{Norm} & \textbf{FFN} & \textbf{Pos-Enc} & \textbf{Window} & \textbf{Params (M)}      \\
\midrule
Whisper & Bidirectional & LayerNorm & GELU     & Sinusoidal  & Fixed & 640 \\
Voxtral Realtime  & Causal  & RMSNorm   & SwiGLU   & RoPE   & Sliding  & 970      \\
\bottomrule
\end{tabular}
\end{table}

\subsection{Adapter Layer}

To reduce the effective sequence length processed by the language decoder, we insert a lightweight adapter layer between the audio encoder and the decoder. This adapter consists of a single MLP that temporally downsamples the encoder outputs, reducing computational cost in the language decoder while preserving relevant acoustic information.

Following Voxtral \citep{liu2025voxtral}, we apply a downsampling factor of 4x, resulting in an effective frame rate of 12.5\,Hz. Therefore, each downsampled audio embedding represents 80\,ms of audio.

\subsection{Language Decoder} \label{sec:lang-dec}

The language decoder is a decoder-only Transformer that operates synchronously with the adapter stream, closely following the delayed-streams decoding scheme of DSM \citep{zeghidour2025streaming}. At each adapter step (80\,ms), the model performs one autoregressive decoding step. The emitted token can be either a text token or a non-emitting placeholder. The placeholder allows the model to ``wait'' when acoustic evidence is insufficient, deferring text emission until the target delay has elapsed. This mechanism enables the model to learn emission timing end-to-end, without external voice activity detection (VAD) or forced alignments. The construction of training targets is described in Section~\ref{sec:training}.

We condition the decoder on a target streaming delay $\tau$, which specifies a minimum offset between acoustic evidence and the earliest time at which corresponding text tokens may be produced, using an Adaptive RMSNorm (AdaRMSNorm) mechanism. $\tau$ is embedded as a sinusoidal embedding and projected using a small MLP with a GELU activation to a vector $g(\tau)\in\mathbb{R}^d$, where $d$ is the model dimension. This conditioning is injected additively in the normalized space on the feed-forward branch of every Transformer block; the attention branch remains unconditioned.

Specifically, given hidden states $x$, a block computes:
\[
r_{\mathrm{attn}} = \mathrm{Attn}(\mathrm{RMSNorm}(x)), \qquad h = x + r_{\mathrm{attn}},
\]
\[
r_{\mathrm{ffn}} = \mathrm{FFN}\!\left(\mathrm{RMSNorm}(h) \odot \left(1.0 + g(\tau)\right)\right), \qquad y = h + r_{\mathrm{ffn}}.
\]
To minimize the additional parameter count, we use an inner-dimension of 32 for the MLP in $g(\tau)$, which introduces 5M extra parameters for the 4.4B model. In Section~\ref{sec:ada-rms}, we show that this form of additive conditioning is more effective than alternative time-conditioning strategies.

The decoder uses sliding window attention with a left-context of 8192 tokens. Together with the encoder’s sliding window, this supports arbitrarily long streams with bounded memory.

%% file: training.tex
\subsection{Target Construction} \label{sec:target-construction}

Training a streaming ASR model requires supervision that aligns a continuous audio stream with a discrete text stream. We leverage $(\text{audio}, \text{text}, \text{word-level timestamps})$ tuples to build frame-synchronous target sequences for the language decoder.

In addition to the base subword vocabulary, we introduce two special symbols: a padding token \texttt{[P]} and a word-boundary token \texttt{[W]}. Training targets are constructed such that the decoder emits exactly one token per downsampled audio frame (80\,ms). For frames in which no text emission is warranted--either because no word is currently underway or because the current word is acoustically incomplete--the target token is \texttt{[P]}. Once a word has been fully observed and the specified target delay has elapsed, a \texttt{[W]} token is emitted to mark the onset of text generation, followed by the subword tokens corresponding to the word itself.

When consecutive words share the same emission frame, no additional \texttt{[W]} token is inserted; the subword tokens of the next word follow directly. We demonstrate in Section~\ref{sec:word-grouping} that this grouping is crucial for retaining the text-modeling capabilities of the language decoder.

This target construction induces an implicit alignment between the audio and text streams. The emission frame of the \texttt{[W]} token defines a grouping point that associates a segment of the audio stream with the subsequent text tokens. The model learns this alignment end-to-end from data, without relying on forced alignments or explicit decoding policies. At inference time, the same learned mechanism governs whether the decoder emits a non-emitting token or initiates text generation at each audio frame.

\subsection{Delay Sampling}

During training, the target delay $\tau$ is sampled uniformly from 80\,ms to 2400\,ms in 80\,ms increments (i.e., 1 to 30 adapter frames). This exposes the model to a range of latency-accuracy tradeoffs, enabling a single model to operate at any delay within this range at inference time via the Ada RMS-Norm conditioning mechanism (Section~\ref{sec:lang-dec}).

\subsection{Optimization}

We initialize the encoder and adapter randomly and the decoder from Ministral 3B \citep{liu2026ministral3}. Training proceeds in two phases:

\begin{enumerate}
    \item \textbf{Encoder warm-up (5\% of training):} The decoder is frozen and only the encoder and adapter are trained. This prevents the randomly initialized encoder from destabilizing the pre-trained decoder representations before it has learned to produce useful audio embeddings.
    \item \textbf{End-to-end (95\% of training):} The full model is trained jointly.
\end{enumerate}

We use the AdamW optimizer \citep{loshchilov2019adamw} with a batch size of 370\,hours. For the encoder warm-up, we use a learning rate of $4 \times 10^{-4}$, and for the end-to-end phase $6 \times 10^{-5}$.

We observed that logit magnitudes in the language decoder grew unboundedly over training. Since we tie the language modeling head and text embedding matrices, this caused text embedding norms to grow proportionally, while audio embedding norms steadily diminished. The resulting imbalance caused the model to increasingly rely on the text stream and ignore audio. We address this by applying a z-loss penalty on the logit norm \citep{debrébisson2016zloss,chowdhery2022palm}, which encourages the softmax normalizer to remain close to zero. This allows the audio and text embedding norms to converge to stable values.

%% file: inference.tex
While low theoretical transcription delay is critical, practical deployments must maintain low latency under realistic conditions (e.g., batching, concurrency, and network overhead). In collaboration with the library authors and community contributors, we contribute realtime serving to the vLLM framework \citep{kwon2023efficient} by combining (i) a paged-attention backend for temporally heterogeneous encoder/decoder KV caches, (ii) resumable streaming sessions that preserve KV state across incremental updates, and (iii) a WebSocket-based realtime endpoint for incremental audio ingestion and token output streaming. Together, these features enable serving Voxtral Realtime with low operational complexity while achieving high throughput.

\subsection{Paged Attention with Temporally Heterogeneous KV Caches}

Voxtral Realtime requires maintaining two KV caches during inference---one for the audio encoder and one for the language decoder---each with a different frame rate. Specifically, the encoder operates at 50\,Hz and the language decoder at 12.5\,Hz, the difference due to $p=4$ temporal pooling applied by the adapter. Thus, one decoder step corresponds to four new encoder KV positions. Standard paged-attention implementations assume a single, uniform KV-position increment per step, which leads to inconsistent block indexing unless the metadata is adapted.

\looseness=-1 To support this efficiently, we implement a custom attention-metadata backend that stretches the encoder-side KV-cache block size by the pooling factor ($p=4$) and applies the same scaling to the associated indexing metadata (sequence lengths and query offsets), while expanding the slot mapping so that each original slot ID maps to a contiguous range of $p$ slots. This keeps vLLM's paged-attention indexing consistent across the encoder and decoder and allows both KV caches to share a unified paged-attention allocation, preserving the performance benefits of vLLM's optimized KV paging.

\subsection{Asynchronous Streaming Input with Resumable Requests}

Most serving frameworks assume the full input is available before decoding begins, which prevents true realtime operation when input arrives continuously. In vLLM, incremental generation is enabled via \emph{resumable requests}: a streaming session persists an anchor request whose KV blocks are reused across incremental updates, so newly arrived input can be appended while reusing previously computed KV states. In our deployment, we pipeline I/O and compute: while the server buffers the next 80\,ms audio increment, it concurrently performs a one-token decoding step, so the next update can be appended and processed immediately when the chunk arrives.

To stream output tokens while audio continues to arrive, vLLM pairs its existing async output generator with a new async input generator, enabling full-duplex streaming (ingest and emit concurrently) rather than a turn-based ``send-then-decode'' loop. A schematic is provided in Appendix~\ref{sec:appendix-b}.

\subsection{WebSocket-Based Realtime API}

To make streaming sessions accessible in production, we contribute a realtime WebSocket API to vLLM. The API provides a bidirectional endpoint for incremental audio ingestion and output token streaming. Clients append audio chunks to an input buffer and periodically commit increments; the server converts these events into resumable session updates for the vLLM engine and streams back token deltas over the same persistent connection with low per-message overhead.

%% file: results.tex
\begin{table}[t]
\centering
\caption{
\textbf{Macro-average WER (\%) across benchmark categories.}. English Short and Long results are averaged across tasks; MCV and FLEURS results are averaged across languages. The definition of "target delay" differs across Realtime APIs and is a function of the audio input. Hence, we omit the delay for the APIs. \textbf{Bold} indicates the best realtime result. \textemdash{} indicates that the multilingual task is unsupported for a mono or bilingual model.}
\label{tab:results}
\begin{tabular}{lccccc}
\toprule
& & \multicolumn{4}{c}{\textbf{WER (\%)}} \\
\cmidrule(lr){3-6}
\textbf{Model} & \textbf{Delay (ms)} & \textbf{En-Short} & \textbf{En-Long} & \textbf{FLEURS} & \textbf{MCV} \\
\midrule

\multicolumn{6}{l}{\textit{Offline}} \\
\cmidrule(lr){1-6}
Whisper & \textemdash{} & 8.39 & 7.97 & 8.23 & 14.25 \\
Voxtral Mini Transcribe V2 & \textemdash{} & 7.27 & 7.11 & 5.90 & 8.07 \\

\addlinespace[0.6em]
\multicolumn{6}{l}{\textit{Realtime API}} \\
\cmidrule(lr){1-6}
GPT-4o mini Transcribe & \textemdash{} & 7.93 & 7.97 & 7.95 & 12.85 \\
Scribe v2 Realtime & \textemdash{} & \textbf{7.33} & 7.43 & 8.34 & 20.85 \\

\addlinespace[0.6em]
\multicolumn{6}{l}{\textit{Realtime Open-Source}} \\
\cmidrule(lr){1-6}
DSM 1B En-Fr & 500 & 12.26 & 13.83 & \textemdash{} & \textemdash{} \\
DSM 2.6B En & 2500 & 8.11 & 7.72 & \textemdash{} & \textemdash{} \\
\addlinespace

\multirow[t]{2}{*}{Nemotron Streaming} & 560  & 9.59 & 14.29 & \textemdash{} & \textemdash{} \\
 & 1120 & 9.41 & 13.02 & \textemdash{} & \textemdash{} \\
\addlinespace

\multirow[t]{4}{*}{Voxtral Realtime} & 240  & 9.95 & 9.29 & 10.80 & 19.22 \\
 & 480  & 8.47 & 7.73 & 8.72 & 15.24 \\
 & 960  & 7.94 & 7.13 & 7.70 & 11.99 \\
 & 2400 & 7.72 & \textbf{6.93} & \textbf{6.73} & \textbf{10.47} \\

\bottomrule
\end{tabular}
\end{table}

We evaluate Voxtral Realtime across English and multilingual benchmarks, comparing against offline systems, realtime APIs, and open-source streaming models. Figure~\ref{fig:fleurs-latency} illustrates the latency--accuracy trade-off of Voxtral Realtime on the FLEURS multilingual benchmark over 13 languages. Full results for each language are presented in Table~\ref{tab:fleurs}. 

At a delay of 480\,ms, Voxtral Realtime approaches the accuracy of Scribe v2 Realtime, the industry-leading realtime API model, as well as Whisper \citep{radford2023whisper}, the most widely adopted offline ASR system. Increasing the delay to 960\,ms further closes the gap, with Voxtral Realtime surpassing past both Scribe v2 Realtime and Whisper. At a higher delay of 2400\,ms, the model continues to improve, achieving accuracy within 1\% of Voxtral Mini Transcribe V2, a state-of-the-art offline transcription model.

Table~\ref{tab:results} reports macro-average WER across four benchmark categories: English short-form, English long-form, FLEURS, and Mozilla Common Voice (MCV). English results are macro-averaged across tasks, while FLEURS and MCV results are averaged across languages. Full results for each task are presented in Appendix~\ref{sec:appendix}.

Across all benchmark categories, Voxtral Realtime substantially outperforms existing open-source streaming baselines at comparable latencies. Prior natively streaming approaches such as DSM achieve competitive accuracy only at substantially higher delays, while Nemotron Streaming \citep{noroozi2024nemotron} exhibits a more limited latency--accuracy trade-off and reduced robustness, particularly in long-form settings. In contrast, Voxtral Realtime consistently improves as latency increases and maintains strong performance across distributions. Notably, Voxtral Realtime supports 13 languages, while other recent open-source models such as DSM support only English and French. 

Taken together, these results demonstrate that Voxtral Realtime achieves offline-level transcription quality at sub-second latency with a natively streaming architecture.

%% file: ablations.tex
In this Section, we ablate three design choices: the delay-conditioning mechanism, the target construction scheme, and the degree of left-padding.

\subsection{Ada RMS-Norm} \label{sec:ada-rms}

There are multiple strategies available to incorporate delay conditioning into the model. DSM sums a sinusoidal delay embedding with the combined audio-text embedding. Alternatively, special tokens can be injected in the text stream to indicate the target delay, though this requires repeating the token at sliding-window boundaries. A third approach using Ada RMS-Norm injects the delay into the decoder's residual stream (Section~\ref{sec:lang-dec}).

\begin{figure*}[t]
    \centering
    \begin{subfigure}[t]{0.32\textwidth}
        \centering
        \includegraphics[width=\linewidth]{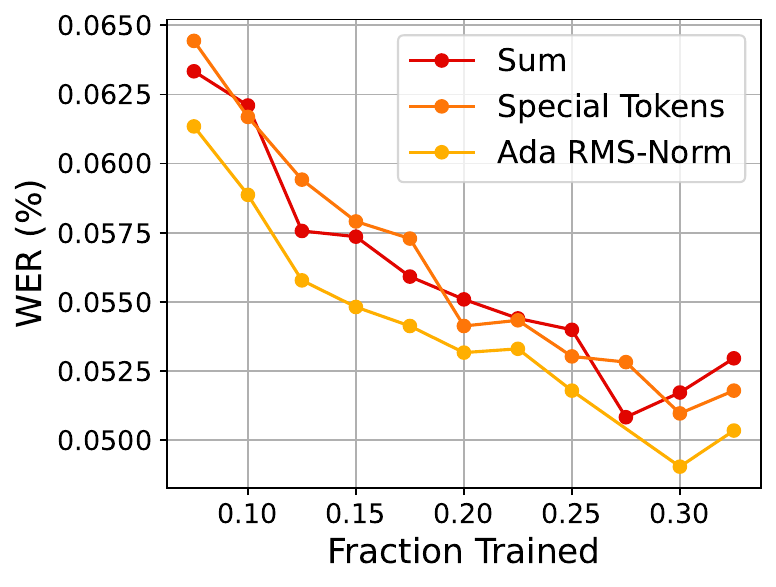}
        \caption{English}
    \end{subfigure}\hfill
    \begin{subfigure}[t]{0.32\textwidth}
        \centering
        \includegraphics[width=\linewidth]{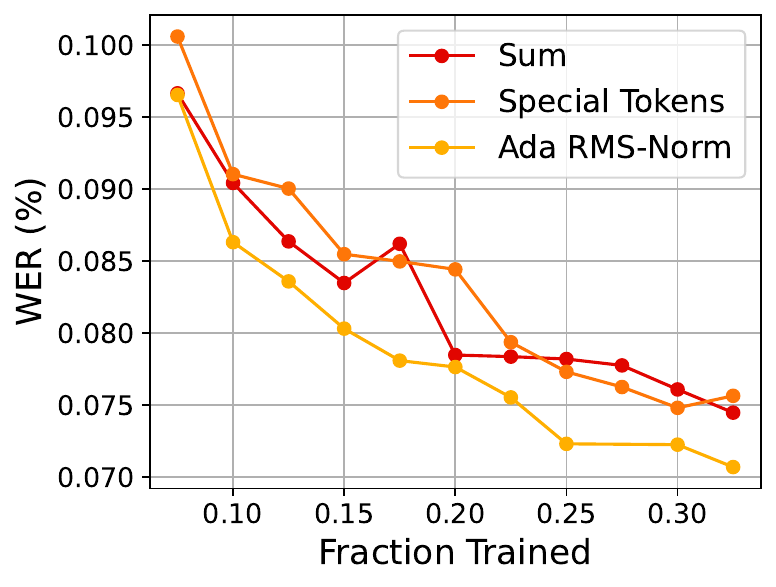}
        \caption{French}
    \end{subfigure}
    \begin{subfigure}[t]{0.32\textwidth}
        \centering
        \includegraphics[width=\linewidth]{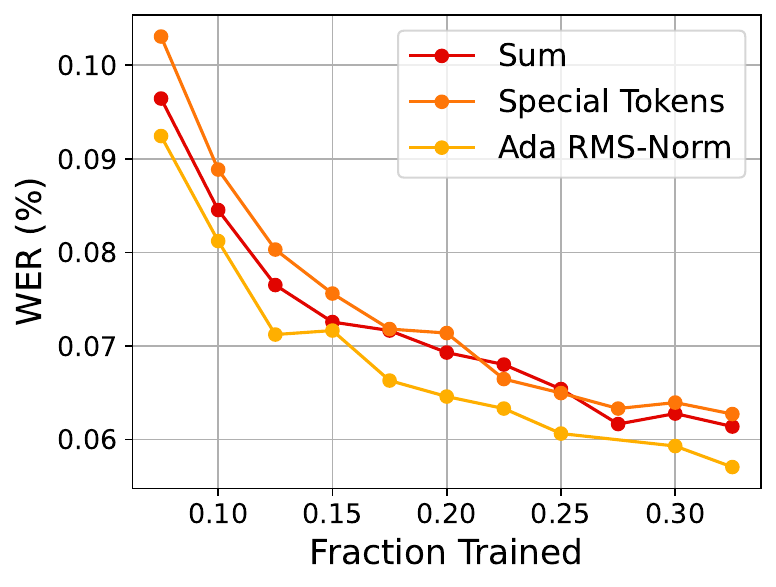}
        \caption{German}
    \end{subfigure}\hfill
    \caption{\textbf{Ablation of delay-conditioning mechanisms.}
    Word error-rate on three languages from the FLEURS dataset as a function of training progress. Ada RMS-Norm consistently improves convergence speed and final accuracy compared to alternative conditioning strategies.}
    \label{fig:abl-delay-cond}
\end{figure*}

Figure~\ref{fig:abl-delay-cond} plots WER results for three languages in the FLEURS dataset for the three conditioning methods outlined above. Summing and special tokens perform comparably, whereas Ada RMS-Norm leads to faster convergence and lower overall WER.

\subsection{Word Grouping} \label{sec:word-grouping}

\begin{figure*}[t]
    \centering
    \begin{subfigure}[t]{0.32\textwidth}
        \centering
        \includegraphics[width=\linewidth]{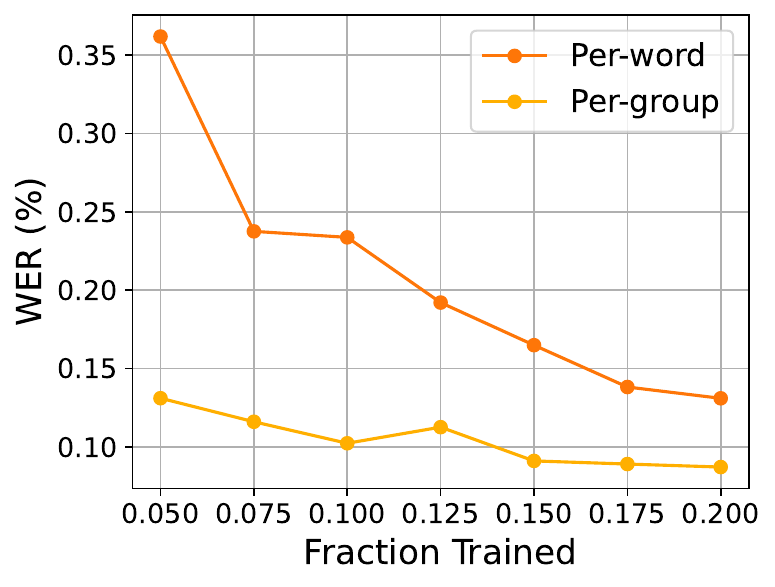}
        \caption{English}
    \end{subfigure}\hfill
    \begin{subfigure}[t]{0.32\textwidth}
        \centering
        \includegraphics[width=\linewidth]{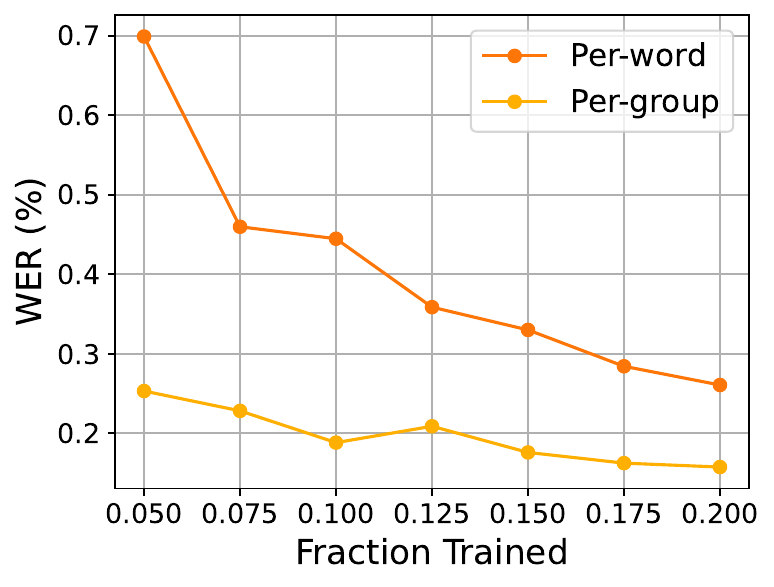}
        \caption{French}
    \end{subfigure}
    \begin{subfigure}[t]{0.32\textwidth}
        \centering
        \includegraphics[width=\linewidth]{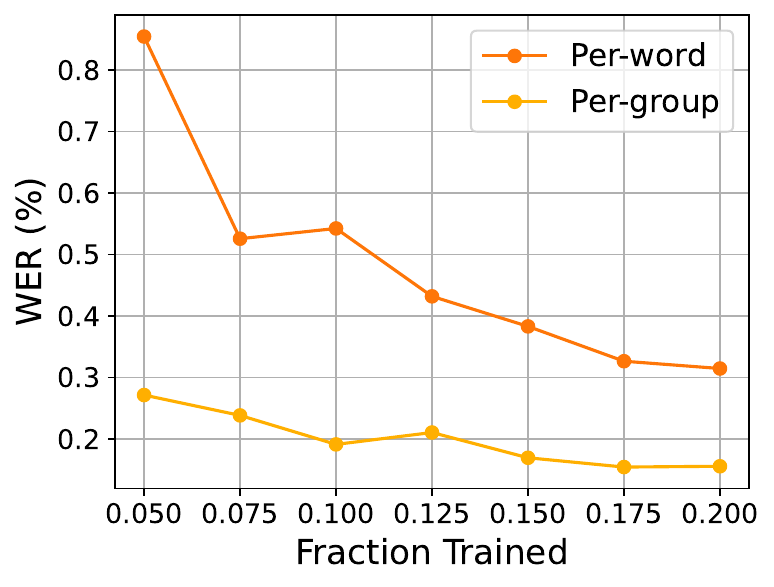}
        \caption{German}
    \end{subfigure}\hfill
    \caption{\textbf{Ablation of target construction schemes.}
    Word error-rate on three languages from the FLEURS dataset as a function of training progress. Inserting a single word-boundary token \texttt{[W]} per-group better preserves the capabilities of the pre-trained language decoder than inserting a \texttt{[W]} per-word.}
    \label{fig:abl-target-construction}
\end{figure*}

In Section~\ref{sec:target-construction}, we describe the construction of the target tokens during training. Figure~\ref{fig:abl-target-construction} compares two target schemes: inserting a word-boundary token \texttt{[W]} between consecutive words in the same emission frame, or grouping them without a boundary. Grouping results in much faster convergence and lower overall WERs. It preserves the subword sequences seen during language model pretraining, allowing the decoder to retain its learned text distributions.

\subsection{Left-Padding}

\looseness=-1 During inference, we explored inserting additional left-padding before the first audio frame. This does not affect streaming delay, as it only increases the size of the prefill step. We pad the audio stream with zeros (equivalent to silence) and the text stream with the corresponding number of \texttt{[P]} tokens.    

Table~\ref{tab:left-padding} shows the WER results across the four benchmark categories as the amount of padding is increased. Increasing the left-padding from 0 to 16 frames improves results across task categories. Further increasing to 32 frames yields additional gains all bar MCV. We hypothesize that left-padding introduces initial tokens that serve a similar role to attention sinks \citep{xiao2024sink}, but leave investigating this to future works.


\begin{table}[t]
\centering
\caption{\textbf{Effect of left-padding on transcription accuracy.} Macro-average WER (\%) across benchmark categories for different degrees of left-padding.}
\label{tab:left-padding}
\begin{tabular}{ccccc}
\toprule
\textbf{Padding Frames} & \textbf{En-Short} & \textbf{En-Long} & \textbf{FLEURS} & \textbf{MCV} \\
\midrule
0 & 9.10 & 10.98 & 9.06 & 16.03 \\
16 & 8.53 & 7.93  & 8.87 & 15.12 \\
32 & 8.47 & 7.73 & 8.72 & 15.24 \\
\bottomrule
\end{tabular}
\end{table}

%% file: conclusion.tex
We introduced Voxtral Realtime, a natively streaming speech transcription model that incorporates a causal encoder, Ada RMS-Norm conditioning, and a training pattern that leverage the pre-trained capabilities of the language decoder.

By achieving near-offline performance at sub-second latency, Voxtral Realtime enables practical real-time applications such as live transcription, voice assistants, and interactive speech interfaces without sacrificing accuracy or language coverage. We release Voxtral Realtime as open weights under the Apache~2.0 license to support further research and deployment of high-quality streaming ASR systems.

%% file: appendix.tex
\begin{landscape}
\section{Appendix}
\subsection{Speech Recognition Results} \label{sec:appendix}

Table \ref{tab:en-short} shows a task-breakdown of short-form English speech recognition results for LibriSpeech Test Clean \citep{panayotov15_librispeech}, LibriSpeech Test Other, GigaSpeech \citep{chen21_gigaspeech}, VoxPopuli \citep{wang21_voxpopuli}, SwitchBoard \citep{godfrey92_switchboard}, CallHome, CHiME-4 \citep{chime4_17}, SPGISpeech \citep{oneill21_spgispeech}, TED-LIUM \citep{Hernandez_2018Tedlium} and Earnings-22 \citep{delrio22_earnings22}. 

\begin{table}[h]
\centering
\caption{English Short-Form WER (\%) results. We report scores for LibriSpeech Test Clean (LS-C), LibriSpeech Test Other (LS-O), GigaSpeech (GS), VoxPopuli (VP), SwitchBoard (SB), CallHome (CH), CHiME-4 (C-4), SPGISPeech (SPGI), TED-LIUM (TED) and Earnings-22 (E22).}
\label{tab:en-short}
\small
\begin{tabular}{lcccccccccccc}
\toprule
\textbf{Model} & \textbf{Delay (ms)} & \textbf{LS-C} & \textbf{LS-O} & \textbf{GS} & \textbf{VP} & \textbf{SB} & \textbf{CH} & \textbf{C-4} & \textbf{SPGI} & \textbf{TED} & \textbf{E22} & \textbf{AVG} \\
\midrule

\multicolumn{13}{l}{\textit{Offline}} \\
\cmidrule(lr){1-13}
Whisper & — & 1.84 & 3.66 & 11.60 & 9.58 & 13.14 & 14.58 & 10.88 & 3.15 & 3.83 & 11.63 & 8.39 \\
Voxtral Mini Transcribe V2 & — & 1.60 & 3.24 & 10.39 & 6.81 & 11.54 & 12.74 & 10.42 & 1.74 & 3.50 & 10.67 & 7.27 \\

\addlinespace[0.6em]
\multicolumn{13}{l}{\textit{Realtime API}} \\
\cmidrule(lr){1-13}
GPT-4o mini Transcribe & — & 1.94 & 4.48 & 10.67 & 6.79 & 11.15 & 15.54 & 11.08 & 2.87 & 4.07 & 10.69 & 7.93 \\
Scribe v2 Realtime & — & 1.75 & 4.01 & 10.29 & 6.01 & 11.54 & 11.87 & 11.59 & 2.04 & 2.63 & 11.52 & 7.33 \\

\addlinespace[0.6em]
\multicolumn{13}{l}{\textit{Realtime Open-Source}} \\
\cmidrule(lr){1-13}
DSM 1B En-Fr & 500 & 3.64 & 11.44 & 12.09 & 11.51 & 12.46 & 16.62 & 28.84 & 4.63 & 4.58 & 16.76 & 12.26 \\
DSM 2.6B En & 2500 & 1.71 & 4.46 & 10.39 & 6.51 & 12.53 & 12.02 & 16.75 & 1.95 & 3.08 & 11.72 & 8.11 \\
\addlinespace
\multirow[t]{2}{*}{Nemotron Streaming} & 560 & 2.42 & 5.12 & 11.87 & 7.09 & 13.82 & 17.30 & 18.64 & 2.64 & 4.50 & 12.48 & 9.59 \\
 & 1120 & 2.38 & 4.94 & 11.84 & 7.02 & 13.35 & 17.74 & 17.37 & 2.62 & 4.50 & 12.33 & 9.41 \\
\addlinespace
\multirow[t]{4}{*}{Voxtral Realtime} & 240 & 2.49 & 7.15 & 12.10 & 10.52 & 13.26 & 14.66 & 18.07 & 3.31 & 4.53 & 13.39 & 9.95 \\
 & 480 & 2.08 & 5.54 & 11.05 & 7.87 & 11.90 & 13.59 & 15.00 & 1.96 & 3.96 & 11.71 & 8.47 \\
 & 960 & 1.96 & 4.59 & 10.51 & 7.23 & 11.44 & 13.34 & 13.17 & 2.36 & 3.55 & 11.24 & 7.94 \\
 & 2400 & 1.82 & 4.03 & 10.60 & 7.06 & 11.55 & 13.44 & 12.18 & 2.11 & 3.57 & 10.80 & 7.72 \\
\bottomrule
\end{tabular}
\end{table}

For English long-form, we report Meanwhile \citep{radford2023whisper} and the long-form version of TED-LIUM. We also take the one-hour long earnings calls from Earnings-21 \citep{delrio21_earnings21} and Earnings-22 \citep{delrio22_earnings22}, and segment them into shorter, 10 minute variants.

\begin{table}[h]
\centering
\caption{English Long-Form WER (\%) results. We report scores for Meanwhile (MW), Earnings-21 (E21), Earnings-22 (E22), and TED-LIUM (TED).}
\label{tab:en-long}
\small
\begin{tabular}{lcccccc}
\toprule
\textbf{Model} & \textbf{Delay (ms)} & \textbf{MW} & \textbf{E21} & \textbf{E22} & \textbf{TED} & \textbf{AVG} \\
\midrule

\multicolumn{7}{l}{\textit{Offline}} \\
\cmidrule(lr){1-7}
Whisper & — & 5.80 & 9.88 & 13.07 & 3.11 & 7.97 \\
Voxtral Mini Transcribe V2 & — & 4.08 & 9.81 & 11.69 & 2.86 & 7.11 \\

\addlinespace[0.6em]
\multicolumn{7}{l}{\textit{Realtime API}} \\
\cmidrule(lr){1-7}
GPT-4o mini Transcribe & — & 5.21 & 9.92 & 12.58 & 4.17 & 7.97 \\
Scribe v2 Realtime & — & 3.62 & 10.72 & 13.22 & 2.18 & 7.43 \\

\addlinespace[0.6em]
\multicolumn{7}{l}{\textit{Realtime Open-Source}} \\
\cmidrule(lr){1-7}
DSM 1B En-Fr & 500 & 7.36 & 14.58 & 21.43 & 11.92 & 13.83 \\
DSM 2.6B En & 2500 & 5.29 & 10.52 & 12.18 & 2.89 & 7.72 \\
\addlinespace
\multirow[t]{2}{*}{Nemotron Streaming} & 560 & 8.25 & 20.92 & 23.53 & 4.46 & 14.29 \\
 & 1120 & 7.43 & 18.75 & 21.65 & 4.25 & 13.02 \\
\addlinespace
\multirow[t]{4}{*}{Voxtral Realtime} & 240 & 5.76 & 12.56 & 14.84 & 4.00 & 9.29 \\
 & 480 & 5.05 & 10.46 & 12.46 & 2.94 & 7.73 \\
 & 960 & 4.14 & 9.86 & 11.63 & 2.86 & 7.13 \\
 & 2400 & 4.03 & 9.52 & 11.31 & 2.86 & 6.93 \\
\bottomrule
\end{tabular}
\end{table}

Tables \ref{tab:fleurs} and \ref{tab:mcv} show the per-language breakdown of error-rate scores for the FLEURS and Mozilla Common Voice benchmarks respectively.

\begin{table}[h]
\centering
\caption{FLEURS error-rate results by language. We report scores for Arabic (ar), German (de), English (en), Spanish (es), French (fr), Hindi (hi), Italian (it), Japanese (ja), Korean (ko), Dutch (nl), Portuguese (pt), Russian (ru), and Chinese (zh). For Chinese and Japanese we report character error-rate (CER). For all other languages, we report WER.}
\label{tab:fleurs}
\small
\begin{tabular}{lccccccccccccccc}
\toprule
\textbf{Model} & \textbf{Delay (ms)} & \textbf{ar} & \textbf{de} & \textbf{en} & \textbf{es} & \textbf{fr} & \textbf{hi} & \textbf{it} & \textbf{ja} & \textbf{ko} & \textbf{nl} & \textbf{pt} & \textbf{ru} & \textbf{zh} & \textbf{AVG} \\
\midrule

\multicolumn{16}{l}{\textit{Offline}} \\
\cmidrule(lr){1-16}
Whisper & — & 15.44 & 5.46 & 4.00 & 2.81 & 5.55 & 28.87 & 2.71 & 4.97 & 14.30 & 5.87 & 3.90 & 5.13 & 7.94 & 8.23 \\
Voxtral Mini Transcribe V2 & — & 13.54 & 3.54 & 3.32 & 2.63 & 4.32 & 10.33 & 2.17 & 4.14 & 12.29 & 4.78 & 3.56 & 4.75 & 7.30 & 5.90 \\

\addlinespace[0.6em]
\multicolumn{16}{l}{\textit{Realtime API}} \\
\cmidrule(lr){1-16}
GPT-4o mini Transcribe & — & 13.99 & 4.07 & 3.65 & 3.41 & 5.84 & 8.39 & 2.82 & 9.89 & 19.46 & 6.00 & 5.04 & 5.30 & 15.43 & 7.95 \\
Scribe v2 Realtime & — & 19.53 & 4.31 & 3.54 & 3.23 & 5.12 & 12.62 & 2.33 & 10.92 & 11.90 & 6.72 & 3.75 & 7.68 & 16.82 & 8.34 \\

\addlinespace[0.6em]
\multicolumn{16}{l}{\textit{Realtime Open-Source}} \\
\cmidrule(lr){1-16}
DSM 1B En-Fr & 500 & — & — & 9.56 & — & 16.31 & — & — & — & — & — & — & — & — & — \\
DSM 2.6B En & 2500 & — & — & 6.11 & — & — & — & — & — & — & — & — & — & — & — \\
\addlinespace
\multirow[t]{2}{*}{Nemotron Streaming} & 560 & — & — & 6.11 & — & — & — & — & — & — & — & — & — & — & — \\
 & 1120 & — & — & 5.72 & — & — & — & — & — & — & — & — & — & — & — \\
\addlinespace
\multirow[t]{4}{*}{Voxtral Realtime} & 240 & 23.95 & 8.15 & 5.91 & 4.59 & 8.00 & 14.26 & 4.41 & 15.17 & 17.56 & 9.23 & 7.51 & 7.87 & 13.84 & 10.80 \\
 & 480 & 22.53 & 6.19 & 4.90 & 3.31 & 6.42 & 12.88 & 3.27 & 9.59 & 15.74 & 7.07 & 5.03 & 6.02 & 10.45 & 8.72 \\
 & 960 & 20.32 & 4.87 & 4.34 & 2.98 & 5.68 & 11.82 & 2.46 & 6.80 & 14.90 & 6.76 & 4.57 & 5.56 & 8.99 & 7.70 \\
 & 2400 & 14.71 & 4.15 & 4.05 & 2.71 & 5.23 & 10.73 & 2.37 & 5.50 & 14.30 & 5.91 & 3.93 & 5.41 & 8.48 & 6.73 \\
\bottomrule
\end{tabular}
\end{table}

\begin{table}[h]
\centering
\caption{Mozilla Common Voice error-rate results by language. For Chinese and Japanese we report CER. For all other languages, we report WER. For fairness, we omit Arabic from the macro-average in Tables~\ref{tab:results} and~\ref{tab:left-padding}, since all models score in excess of 45\%.}
\label{tab:mcv}
\small
\begin{tabular}{lccccccccccccccc}
\toprule
\textbf{Model} & \textbf{Delay (ms)} & \textbf{ar} & \textbf{de} & \textbf{en} & \textbf{es} & \textbf{fr} & \textbf{hi} & \textbf{it} & \textbf{ja} & \textbf{ko} & \textbf{nl} & \textbf{pt} & \textbf{ru} & \textbf{zh} & \textbf{AVG} \\
\midrule

\multicolumn{16}{l}{\textit{Offline}} \\
\cmidrule(lr){1-16}
Whisper & — & 50.58 & 6.25 & 22.91 & 5.66 & 11.33 & 46.75 & 6.81 & 15.80 & 20.86 & 5.83 & 7.17 & 6.76 & 14.88 & 14.25 \\
Voxtral Mini Transcribe V2 & — & 46.06 & 4.35 & 8.61 & 3.93 & 7.21 & 10.26 & 4.15 & 12.87 & 20.29 & 4.38 & 6.58 & 5.18 & 9.04 & 8.07 \\

\addlinespace[0.6em]
\multicolumn{16}{l}{\textit{Realtime API}} \\
\cmidrule(lr){1-16}
GPT-4o mini Transcribe & — & 51.06 & 6.05 & 10.89 & 5.54 & 9.77 & 23.90 & 5.75 & 18.53 & 32.90 & 7.89 & 9.70 & 8.49 & 14.81 & 12.85 \\
Scribe v2 Realtime & — & 60.60 & 16.60 & 19.43 & 15.93 & 15.74 & 35.78 & 14.04 & 24.70 & 26.98 & 9.06 & 19.77 & 13.16 & 38.97 & 20.85 \\

\addlinespace[0.6em]
\multicolumn{16}{l}{\textit{Realtime Open-Source}} \\
\cmidrule(lr){1-16}
DSM 1B En-Fr & 500 & — & — & 34.93 & — & 24.29 & — & — & — & — & — & — & — & — & — \\
DSM 2.6B En & 2500 & — & — & 18.27 & — & — & — & — & — & — & — & — & — & — & — \\
\addlinespace
\multirow[t]{2}{*}{Nemotron Streaming} & 560 & — & — & 12.33 & — & — & — & — & — & — & — & — & — & — & — \\
 & 1120 & — & — & 11.92 & — & — & — & — & — & — & — & — & — & — & — \\
\addlinespace
\multirow[t]{4}{*}{Voxtral Realtime} & 240 & 55.10 & 11.13 & 19.63 & 9.05 & 14.51 & 20.05 & 10.62 & 27.25 & 33.47 & 11.86 & 15.27 & 13.90 & 43.93 & 19.22 \\
 & 480 & 48.64 & 8.70 & 15.18 & 6.05 & 11.51 & 17.22 & 7.80 & 20.87 & 31.37 & 8.97 & 11.25 & 11.03 & 32.92 & 15.24 \\
 & 960 & 48.68 & 6.85 & 12.49 & 5.12 & 9.80 & 15.04 & 6.05 & 16.77 & 27.24 & 6.36 & 8.00 & 8.65 & 21.51 & 11.99 \\
 & 2400 & 50.35 & 5.66 & 10.51 & 4.56 & 9.05 & 13.19 & 4.96 & 15.45 & 25.26 & 5.18 & 7.22 & 7.64 & 17.00 & 10.47 \\
\bottomrule
\end{tabular}
\end{table}

\end{landscape}

\subsection{vLLM Realtime Inference} \label{sec:appendix-b}
\begin{figure}[h]
    \centering
    \includegraphics[width=\textwidth]{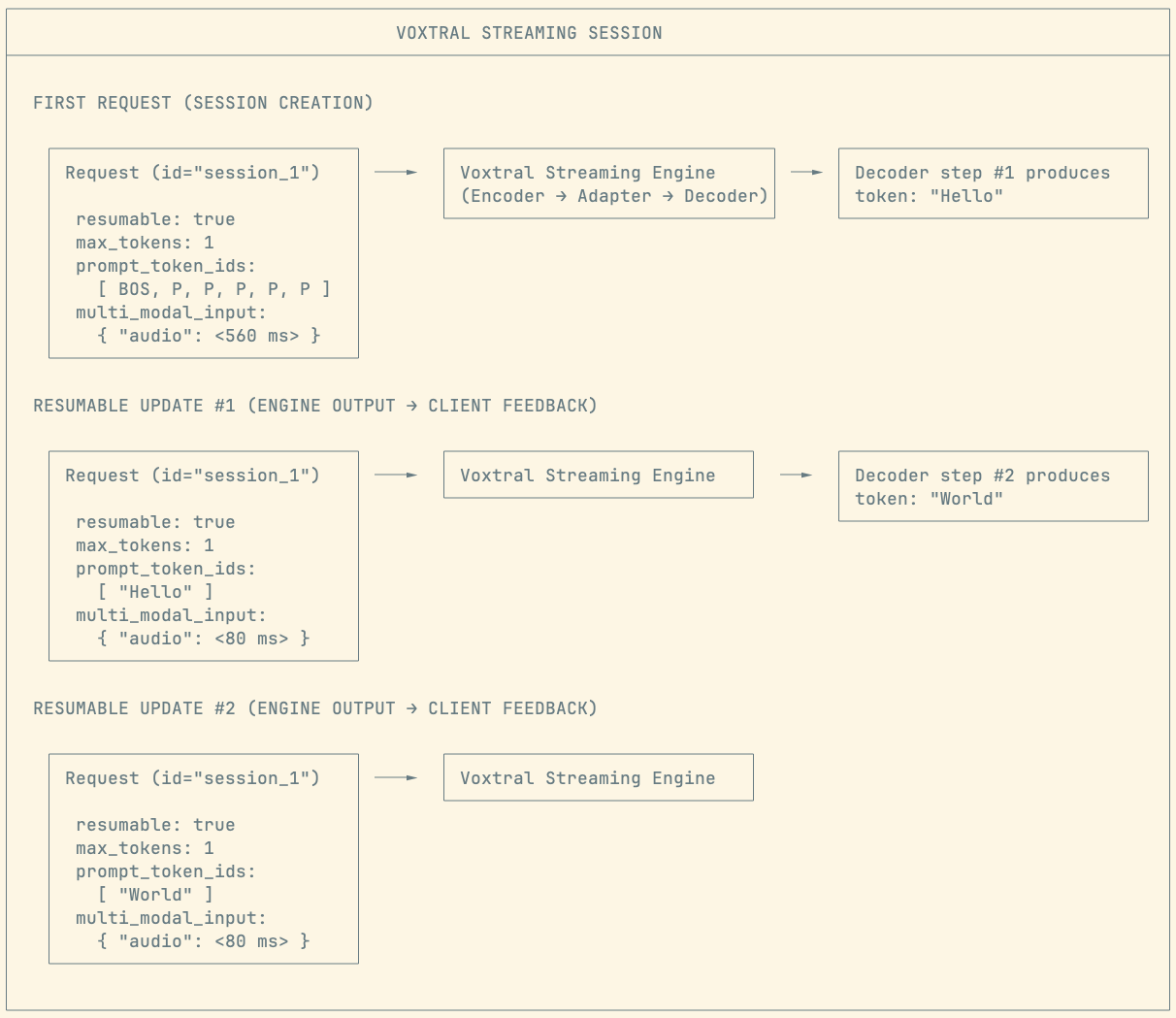}
    \caption{\textbf{Voxtral streaming session via vLLM resumable requests.} A session is created with an anchor request that includes the initial buffered audio (e.g., the first $\tau$\,ms plus padding tokens to enforce the target delay) and runs a one-token decoder step. Each subsequent update is sent as a resumable request that appends the next 80\,ms audio chunk together with the previously emitted token ID, allowing the engine to reuse cached KV states and emit the next token incrementally. This request--decode--update loop enables low-latency, continuous transcription with full-duplex streaming-input/streaming-output.}
    \label{fig:vllm-session}
\end{figure}